\documentclass{article}
\usepackage{ijcai25}

\usepackage{times}
\usepackage{soul}
\usepackage{url}
\usepackage[hidelinks]{hyperref}
\usepackage[utf8]{inputenc}
\usepackage[small]{caption}
\usepackage{graphicx}
\usepackage{amsmath}
\usepackage{amsthm}
\usepackage{booktabs}
\usepackage{algorithm}
\usepackage{algorithmic}
\usepackage[switch]{lineno}
\usepackage{textcomp}
\usepackage{epstopdf}
\usepackage{booktabs}
\usepackage{amsmath}
\usepackage{hyperref}
\usepackage{tipa}
\usepackage{esint}
\usepackage{arydshln}
\usepackage{multirow}

\usepackage{amssymb}

\usepackage{color}
\usepackage{xcolor}
\usepackage{utfsym}
\usepackage{colortbl}
\usepackage[shortcuts]{extdash}
\usepackage{stfloats}
\usepackage{float}
\usepackage{makecell}
\usepackage{newtxmath} 
\usepackage{caption}
\usepackage{colortbl}

\usepackage{amsmath}
\usepackage{bm}
\usepackage{amsfonts}
\usepackage{amsmath}
\usepackage{bigstrut}

\usepackage{xspace}

\newcommand{\methodname}{UPC\xspace}

\def\hjw{\textcolor{black}}

\def\c{\textcolor{black}}
\def\cam{\textcolor{black}}

\title{Enhancing User-Oriented Proactivity in Open-Domain Dialogues \\ with Critic Guidance}

\author{
Yufeng Wang$^{1,2,3}$\thanks{Equal contribution. $^\dagger$Corresponding author. Code and Appendix are available at \href{https://github.com/wang678/LLM-UPC}{https://github.com/wang678/LLM-UPC}.}
\and
Jinwu Hu$^{1,3*}$\and
Ziteng Huang$^{1}$\and
Kunyang Lin$^{4}$\and
Zitian Zhang$^{1}$\and \\
Peihao Chen$^{5}$\and
Yu Hu$^{6}$\and
Qianyue Wang$^{1}$\and
Zhuliang Yu$^{1}$\and
Bin Sun$^{7}$\and
Xiaofen Xing$^{1\dagger}$\and \\
Qingfang Zheng$^{2\dagger}$\and
Mingkui Tan$^{1,3}$
\vspace{4pt}
\affiliations
$^1$South China University of Technology
$^2$Peng Cheng Laboratory
$^3$Pazhou Laboratory \\
$^4$Tencent AI Lab
$^5$Tencent Robotics X Lab
$^6$Hong Kong Polytechnic University
$^7$Hunan University\\
\emails
yufeng6568@gmail.com, xfxing@scut.edu.cn, zhengqf01@pcl.ac.cn
}

\begin{document}

\maketitle

\begin{abstract}
Open-domain dialogue systems aim to generate natural and engaging conversations, providing significant practical value in real applications such as social robotics and personal assistants. The advent of large language models (LLMs) has greatly advanced this field by improving context understanding and conversational fluency. However, existing LLM-based dialogue systems often fall short in proactively understanding the user's chatting preferences and guiding conversations toward user-centered topics. This lack of user-oriented proactivity can lead users to feel unappreciated, reducing their satisfaction and willingness to continue the conversation in human-computer interactions. To address this issue, we propose a \textbf{U}ser-oriented \textbf{P}roactive \textbf{C}hatbot (\textbf{\methodname}) to enhance the user-oriented proactivity. Specifically, we first construct a critic to evaluate this proactivity inspired by the LLM-as-a-judge strategy. Given the scarcity of high-quality training data, we then employ the critic to guide dialogues between the chatbot and user agents, generating a corpus with enhanced user-oriented proactivity. To ensure the diversity of the user backgrounds, we introduce the \textit{ISCO-800}, a diverse user background dataset for constructing user agents. Moreover, considering the communication difficulty varies among users, we propose an iterative curriculum learning method that trains the chatbot from easy-to-communicate users to more challenging ones, thereby gradually enhancing its performance. Experiments demonstrate that our proposed training method is applicable to different LLMs, improving user-oriented proactivity and attractiveness in open-domain dialogues. 
\end{abstract}

\section{Introduction}


Humans are naturally more engaged in conversations relevant to their interests~\cite{oertel2020engagement}. To meet humans' chatting preferences, the open-domain dialogue task aims to build chatbots that produce engaging and meaningful natural language responses \cite{kann2022open}. It yields practical value in various human-computer interaction applications, such as personal assistants~\cite{luo2022critical}, education~\cite{rodrigues2022review} and social robotics~\cite{grassi2022knowledge}. 
\c{Recently, the rise of large language models (LLMs)~\cite{zhao2023survey,hu2024dynamic} has led to unprecedented popularity in dialogue systems. LLM-based dialogue systems have demonstrated extraordinary context understanding and coherent response generation capabilities~\cite{deng2023rethinking}, significantly advancing the development of open-domain dialogue.}

\begin{figure*}[t]
\vspace{-10pt}
\centering
\includegraphics[width=0.78\linewidth]{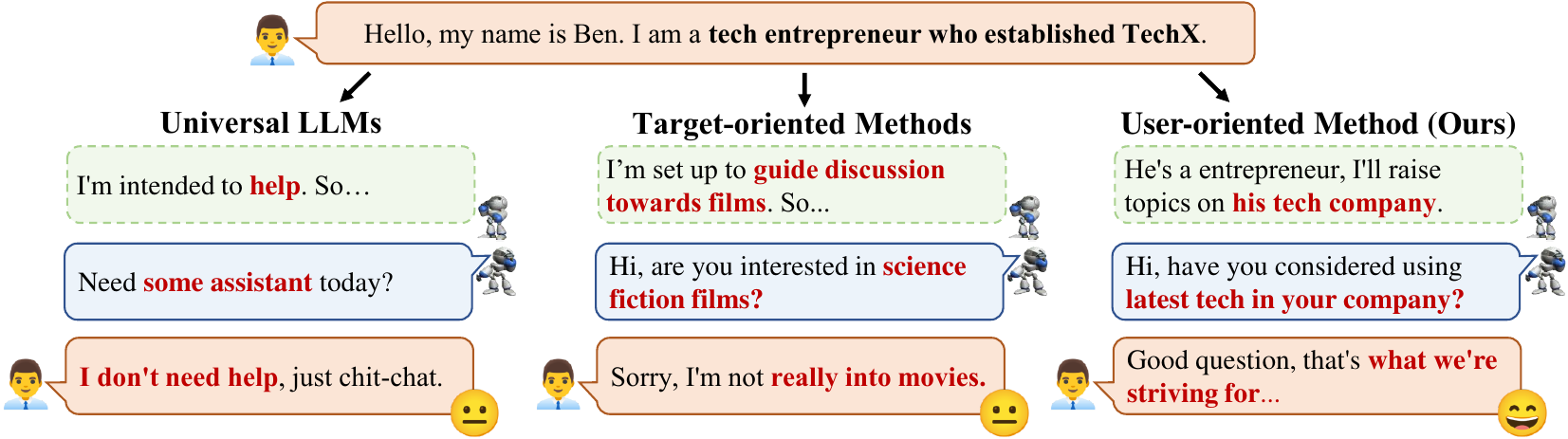}
\vspace{-8pt}
    \caption{
    Illustration of different open-domain dialogue methods. Left: universal LLMs. Medium: target-oriented methods. Right: our user-oriented method. The green dotted box means the intention of the chatbot.
}  
    \label{fig:teaser}
\vspace{-12pt}
\end{figure*}

Despite excellent dialogue capabilities of LLMs, they still have limitations due to the inability of \textit{proactivity}.
\c{The proactive chatbot is designed to go beyond passive response to user inputs, but actively explore the user's chatting interests and lead the topics to the user's preferences.}
Since LLM-based methods heavily rely on the existing training conversations and knowledge to provide suggestions and complete tasks~\cite{openai2023gpt}, they typically generate responses to user queries in a passive manner~\cite{liao2023proactive}. They tend to help solve problems rather than proactively learn about the background and chatting interests of the user (Figure \ref{fig:teaser}, left). 
Existing methods enhance proactivity by topic planning and shifting, or topic-aware response generation~\cite{tang2019target,deng2023survey,deng2023prompting,wang2024target}. 
Although they can actively lead the chatting topic, these target-oriented methods focus more on target-centered topic switching,  resulting in a lack of proactive attention toward the users themselves (Figure \ref{fig:teaser}, medium). This may not fully align with the original intention of open-domain chatbots, which is to enhance the user's chat experience and create engaging dialogues. \c{To address this, we \textbf{delve into the proactivity of exploring user's background and chatting interests and take proactivity to lead the conversation toward user-centered topics}, namely the} \textit{User-oriented Proactivity} (Figure \ref{fig:teaser}, right). It enables the chatbot to deliver responses that are not only relevant but also resonate with the user's chatting interests and background, thereby fostering more satisfying interaction.

In this paper, we propose a \textbf{U}ser-oriented \textbf{P}roactive \textbf{C}hatbot (\textbf{\methodname}) to address the lack of user-oriented proactivity. We build a chatbot that explores the user’s background and chatting interests, and actively leads the conversation towards user-centered topics. Specifically,
since assessing proactivity by humans is costly, LLM can serve as a judge that aligns with humans~\cite{zheng2024judging}. We appoint an LLM to be a critic and design evaluation metrics for this proactivity. 
Besides, the lack of a high-quality training corpus for a proactive chatbot puts us in a dilemma of making bricks without straw. To this end, 
we introduce a critic-guided dialogue corpus generation paradigm. We employ the critic to guide dialogues between the chatbot and different user agents, generating a dialogue corpus with enhanced user-oriented proactivity.
Since communication difficulty varies across users, the chatbot may struggle with users who have uncommon backgrounds or preferences in the early training stages. Therefore,
we propose a communication difficulty-aware iterative curriculum learning approach. It adapts the chatbot to easy-to-communicate users in early training iterations and incrementally learns the chatbot with more challenging users, achieving steady performance improvement.
Finally, we construct the \textit{ISCO-800}, a dataset with 800 user backgrounds, to create diverse user agents.
Our main contributions are as follows: 

\textbf{1) Design of a critic to assess user-oriented proactivity.} To address the lack of user-oriented proactivity in the chatbot, we develop a critic to quantify and assess this proactivity. It also guides high-quality corpus generation for training. The real user evaluation verifies its alignment with humans.

\textbf{2) Iterative curriculum learning is proposed for chatbot training.} \c{Recognizing that communication difficulty varies among different users, we define adaptation difficulty and introduce an iterative curriculum learning method. It first adapts the chatbot to easy-to-communicate users in early training iterations and progressively includes more challenging ones. Experiments show that this paradigm improves average performance by 8.2\% compared to the original LLM.}

\textbf{3) Construction of a user background dataset \textit{ISCO-800}.} To train a chatbot capable of interacting with users from various backgrounds. We construct a user background dataset that includes 800 types of user background information. It provides realistic background information, including occupation, hobby, education, and personality, enabling the chatbot to interact with a wide range of users.

\section{Related Works}
\label{related work}
Significant efforts have been made to enhance the human-like performance of open-domain dialogue systems. Based on key issues, existing studies relevant to our work can be broadly categorized into three areas: coherent dialogue, personalized dialogue, and target-oriented dialogue. 

\textbf{Coherent dialogue} aims to improve contextual coherence during conversations. Memochat~\cite{lu2023memochat} uses memorization-retrieval-response cycles to teach the LLMs to memorize and retrieve past dialogues with structured memos, leading to enhanced consistency. 
Sun \textit{et al.}~\cite{sun2023towards} propose a hybrid latent variable method. It combines discrete and continuous latent variables to improve both semantic correlation and diversity. Although these methods improve coherence, they do not prioritize the proactive, user-centric dialogue that is central to our work.

\textbf{Personalized dialogue} focuses on building a chatbot endowed with a persona~\cite{tu2023characterchat,chen2023towards}, or modeling the persona of the other party~\cite{chen2024recent,zhong2024memorybank}. For example, ORIG~\cite{chen2023towards} proposes a model-agnostic framework to fine-tune the persona dialogue model. It enables the dialogue models to learn robust representations and improve the consistency of response generation.
In addition to improving the chatbot's role-playing abilities, Memorybank~\cite{zhong2024memorybank} adds long-term memory functionality for LLMs by summarizing past interactions to capture user personalities. However, unlike the static personality, the user's chatting interests evolve dynamically during a conversation. These methods often struggle to actively explore and engage with these changing interests, limiting their ability to generate topics that align with what the user is currently interested in.

\textbf{Target-oriented dialogue} improve proactivity by planning and steering the dialogue to pre-set topics~\cite{tang2019target,deng2023survey,deng2023prompting,wang2024target}. 
\c{For example, TRIP~\cite{wang2024target} generates the dialogue path with a bidirectional planning method to drive the conversation to the goal. ProCoT~\cite{deng2023prompting} proposes the proactive chain-of-thought prompting scheme, which augments LLMs with the goal planning capability.} 
Although they actively lead the conversation towards system-side goals, they lack the ability to understand the user's chatting interests, resulting in low engagement and a diminished willingness to continue the conversation. In contrast, our method emphasizes user-oriented proactivity by focusing on topics that are centered around the user's background and interests.

\begin{figure*}[t]
\vspace{-10pt}
\centering
\includegraphics[width=0.9\linewidth]{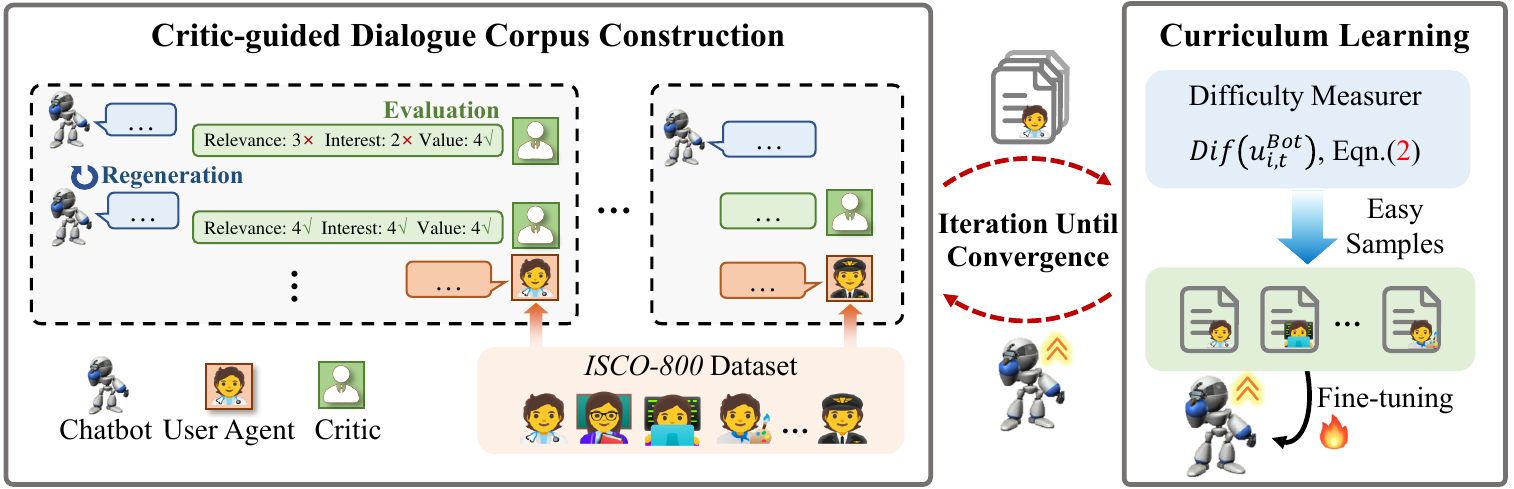}
\vspace{-5pt}
    \caption{
    General diagram of proposed \methodname. We train \methodname~in an iterative process. In each iteration, the chatbot engages in dialogue with user agents from the \textit{ISCO-800} dataset, guided by the critic to generate a corpus with enhanced proactivity. In the iterative curriculum learning process, we measure the communication difficulty and fine-tune the chatbot with the corpus corresponding to easy-to-communicate users. The fine-tuned chatbot is then used to generate the corpus for the next iteration. This iterative process repeats until convergence. 
    }  
    \label{fig:method}
\vspace{-5pt}
\end{figure*}

\section{Problem Definition}
Given a chatbot $\mathcal{C_\theta}$ parameterized by $\theta$, we consider the task of having a multi-turn open-domain conversation between the chatbot and each user $\mathcal{U}_i$ from a user set $\{\mathcal{U}_i\}_{i=1}^N$. The resulting dialogue corpus is denoted as $\{\mathcal{D}_i\}_{i=1}^N $ with the total dialogue number of $N$.  $\mathcal{D}_i=\left\{ \left( u_{i,t}^{Bot}, u_{i,t}^{User} \right)_{t=1}^T  \right\}$ is the one conversation with $T$ turns, and $u_{i,t}^{Bot}$, $u_{i,t}^{User}$ are the utterances of the chatbot and user at turn $t$, respectively.

Existing methods often passively respond to the user queries or focus on target-oriented topics. This lack of user-oriented proactivity can make users feel unappreciated, reducing their satisfaction and willingness to continue the conversation in human-computer interactions. 
\c{Therefore, our goal is to develop a chatbot that proactively explores the user's chatting interests and guides conversations toward user-centered topics. To evaluate this kind of proactivity, the primary evaluation metric is whether the user finds the chatbot's response interesting. In addition, background relevance plays a crucial role, as the user typically focuses on topics related to their context. Finally, the value of the chatbot's response is an important factor, as it directly influences the quality of the conversation. To this end,
We aim to build a chatbot with user-oriented proactivity from the perspective of the user's chatting interests $S_{int.}^{i,t}$,  background relevance $S_{rel.}^{i,t}$ and response value $S_{val.}^{i,t}$, respectively. 
These metrics gauge the chatbot's ability to proactively explore the user's background, engage with the user's desired topics, and provide responses that are both interesting and valuable. }

\section{User-Oriented Proactive Chatbot}
The overall framework for \methodname~training is shown in Figure \ref{fig:method}. We first construct a critic to evaluate user-oriented proactivity, then train the chatbot in an iterative curriculum learning process. 
In each iteration, the chatbot engages in a dialogue with different user agents, each with a distinct background from our \textit{ISCO-800} dataset. During the conversation, the chatbot receives evaluations from the critic, which guides the generation of a high-quality corpus with improved proactivity.
Since the communication difficulty varies across users, we assess the difficulty based on the quality of the generated corpus, so as to fine-tune the chatbot with the corpus corresponding to easy users. Then, the fine-tuned chatbot is used to start a new iteration of dialogue corpus generation. 
We repeat this iterative process to improve the performance until convergence.

\begin{figure}[t]
\vspace{-5pt}
\centering
\includegraphics[width=0.9\linewidth]{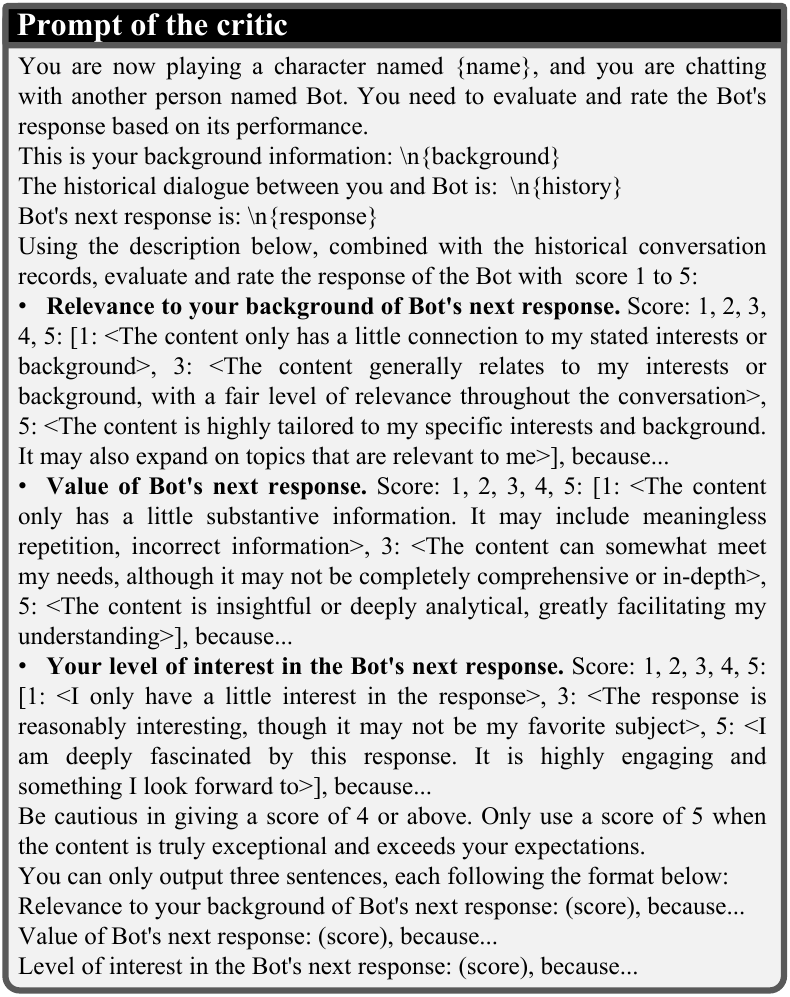}
\vspace{-5pt}
    \caption{
    The prompt of the critic.
    }  
    \label{fig:critic_prompt}
\vspace{-11.5pt}
\end{figure}

\subsection{Evaluation of User-oriented Proactivity}
\label{sec:Metric}
To evaluate the user-oriented proactivity $\mathbf{S}_{i,t}$ from the three perspectives of user's interest level, background relevance, and response value, we follow the LLM-as-a-judge method~\cite{zheng2024judging} and construct the critic $\mathcal{J}$ by prompting the ChatGPT to score the abilities. Specifically, we let the critic obtain the user's background and the dialogue history $\mathcal{D}_{i,t-1}$ between the user and the chatbot. Then we use the 5-point scoring system to evaluate the current response $u_{i,t}^{Bot}$ of the chatbot. It is formalized:
\begin{equation}
\label{eq:eval}
\mathbf{S}_{i,t}=\left[S_{int.}^{i,t}, S_{rel.}^{i,t}, S_{val.}^{i,t}\right] = \mathcal{J}\left( u_{i,t}^{Bot}, \mathcal{U}_i, \mathcal{D}_{i,t-1} \right),
\end{equation}
where the dialogue history $\mathcal{D}_{i,t-1}=\left\{ \left( u_{i,j}^{Bot}, u_{i,j}^{User} \right)_{j=1}^{t-1}  \right\}$. 
Figure \ref{fig:critic_prompt} illustrates the prompt of critic. We provide descriptions for scores of 1, 3, and 5 to improve the rating accuracy.

\begin{figure}[t]
\centering
\includegraphics[width=0.9\linewidth]{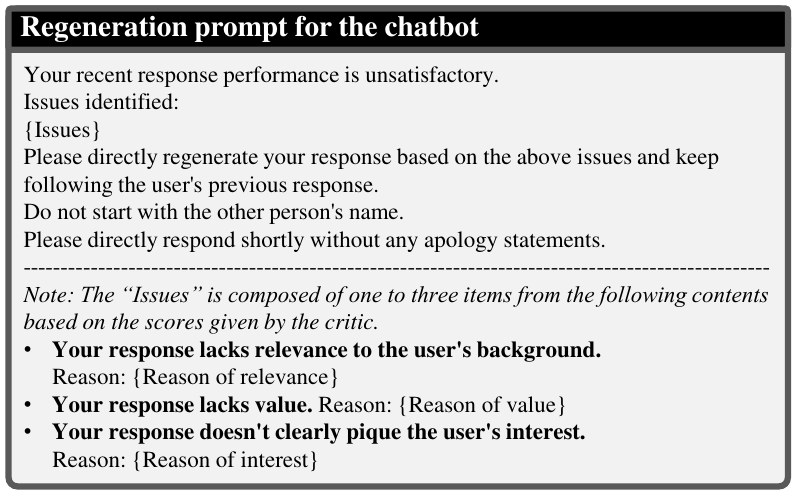}
\vspace{-8pt}
    \caption{
    The regeneration prompt for the chatbot.
    }  
    \label{fig:regen_prompt}
\vspace{-10pt}
\end{figure}

\subsection{Critic-guided Dialogue Corpus Generation}
\label{subsec: corpus_collection}
\c{We employ the critic to guide dialogues between the chatbot and user agents, generating a dialogue corpus with enhanced user-oriented proactivity.} We use Qwen1.5-72B-Chat, an LLM with strong role-playing abilities~\cite{Chatbot_Arena}, to act as user agents. Each user agent is assigned a unique background from our \textit{ISCO-800} dataset, described in Section \ref{sec:isco800}. The system prompt of the user agent is shown in Appendix E. We then put the chatbot and each user agent into multiple rounds of open-domain conversation. Notably, the chatbot does not have prior access to the user's background but is instead prompted to actively learn about the background of the user during the conversation, and find topics of potential interest for the user. The system prompt is shown in Appendix E. In the first round, we prompt the chatbot to greet first and let the user give a brief self-introduction. Then they engaged in open-ended conversation. The critic evaluates each response from the chatbot, excluding the initial greeting. Each response is scored on three dimensions: user interest, background relevance, and response value. Scores range from 1 to 5, with corresponding reasons provided. For responses scoring below 4, the chatbot is prompted to regenerate based on the critic's feedback, as shown in Figure \ref{fig:regen_prompt}. The regeneration continues until all scores are 4 or higher, or until the maximum number of regenerations is reached.
Our critic-guided approach ensures that the chatbot generates high-quality responses by trial and error like humans~\cite{young2009learning}, progressively enhancing user-oriented proactivity. Besides, the proportion of high-quality responses increases with the evolution of the chatbot. The corpus generation pipeline in each training iteration $k$ is detailed in Algorithm \ref{alg:datacollect_algorithm}.

\begin{algorithm}[t]

\footnotesize
    \caption{Dialogue corpus generation  in iteration $k$.}
    \label{alg:datacollect_algorithm}
    \begin{algorithmic}[1]
    \vspace{-1pt}
    \REQUIRE
    The chatbot $\mathcal{C}_{\theta_{k}}$ in iteration $k$, the user agent $\mathcal{U}_i$ from \textit{ISCO-800} dataset $\{\mathcal{U}_i\}_{i=1}^N$.  The critic $\mathcal{J}$. The maximum re-gen attempts $R$, dialogue turns $T$ and score buffer $\mathcal{S}_i$. 
    \FOR{$i = 1,..., N$}
        \FOR{$t = 1,..., T$}
            \IF {$t == 1$}
                \STATE The chatbot $\mathcal{C}_{\theta_{k}}$ gives an greeting utterance $u_{i,t}^{Bot}$. 
                \STATE The user agent $\mathcal{U}_i$ gives brief self-introduction $u_{i,t}^{User}$. 
                \STATE Add $(u_{i,t}^{Bot}, u_{i,t}^{User})$ to the dialogue history $\mathcal{D}_{i,t}$.

            \ELSE
                \STATE The chatbot $\mathcal{C}_{\theta_{k}}$ gives response $u_{i,t}^{Bot}$.

                \STATE Compute scores $\mathbf{S}_{i,t}$ for current  $u_{i,t}^{Bot}$ via Eqn. (\ref{eq:eval}).
                \STATE Cache scores, utterances: $\mathbf{S}^{\prime}_{i,t} \leftarrow \mathbf{S}_{i,t}$, $u_{i,t}^{\prime Bot} \leftarrow u_{i,t}^{Bot}$.
                \STATE Initialize $r = 0$.
                \WHILE{$(S_{int.}^{i,t}$, $S_{rel.}^{i,t}$ or $S_{rel.}^{i,t}<4)~and~r <= R$}
                    \STATE Re-generate utterance $u_{i,t}^{Bot}$, $r = r + 1$.
                    \STATE Compute scores $\mathbf{S}_{i,t}$ for $u_{i,t}^{Bot}$ via Eqn. (\ref{eq:eval}).
                \ENDWHILE
                \STATE The user agent $\mathcal{U}_i$ gives response $u_{i,t}^{User}$.
                \STATE Add $(u_{i,t}^{\prime Bot},u_{i,t}^{Bot}, u_{i,t}^{User})$ to $\mathcal{D}_i$, Add $(\mathbf{S}^{\prime}_{i,t}, \mathbf{S}_{i,t})$ to $\mathcal{S}_i$.

            \ENDIF
        \ENDFOR
    \ENDFOR

    \RETURN Dialogue corpus  $\{\mathcal{D}_i\}_{i=1}^N $ and scores $\{\mathcal{S}_i\}_{i=1}^N $  in iter. $k$.

    \end{algorithmic}
    \vspace{-2pt}

\end{algorithm}

\subsection{Communication Difficulty-Aware Iterative Curriculum Learning}
Since the training data is generated through interactions between the chatbot and the user, timely improvement of the chatbot helps enhance the generated data and final performance. Therefore, it is crucial to conduct iterative training that uses the model from the previous iteration to generate data for the next iteration. Besides, since the communication difficulty varies among different users~\cite{lindqvist2015interprofessional},
similar to human learning, fine-tuning models benefit from an easy-to-difficult curriculum during model training \cite{gao2024confucius,hu2025dynamic}. Therefore, we propose a communication difficulty-aware iterative curriculum learning framework that consists of a Communication Difficulty Measurer and Training Scheduler.

\hjw{\textbf{Communication Difficulty Measurer.} \cam{Our difficulty
measurer captures two aspects: how well the chatbot performs
and whether it can improve. The first checks if the chatbot can
produce acceptable responses for the user. The second evaluates whether the chatbot can make meaningful progress from
the feedback. Therefore,} we use the scores described in Section \ref{sec:Metric} and their improvement with the critic as the Difficulty Measurer ${Dif}(u_{i,t}^{Bot})$}:
\begin{equation}
\label{equation: diff}
{Dif}(u_{i,t}^{Bot}) = True \text{~~if~~} P \text{~~else~~} False,
\end{equation}
the condition $P$ is formalized:
\begin{equation}
    \label{equation: condition}
    P:\exists p \in \{ S_{int.}^{i,t},S_{rel.}^{i,t},S_{val.}^{i,t} \} \ s.t. \ p \textless \alpha \ or  \sum_{c \in \{\mathbf{S}_{i,t}-\mathbf{S}_{i,t}^\prime \}} \mathbb{I}(c \textgreater 0) \textless \ \beta ,
\end{equation}
where $True$ represents difficult and the reverse is easy. $\alpha$ and $\beta$ are hyperparameters, specified in Appendix B. A sample is considered difficult if any metric is below $\alpha$ or the number of boosted metrics is fewer than $\beta$.  \cam{These criteria identify user samples that are either hard to generate high-quality responses or hard to improve.}

\hjw{\textbf{Training Scheduler.} We adjust the generated training corpus at each training iteration $k$ with the Difficulty Measurer for the collected multi-round dialogue corpus. Specifically, for the dialogue corpus $\mathcal{D}$ \c{of different users}, we follow Eqn. (\ref{equation: diff}) to measure the difficulty of the \c{corpus $\{\mathcal{D}_i\}_{i=1}^N $}. The easy corpus constitutes the training dataset $\mathcal{D}^*$, while the difficult corpus will be discarded and its corresponding users will be engaged in dialogues in the next round. Notably, as the chatbot is updated, its ability to adapt to a wider range of users improves, leading to an increase in the amount of easy corpus. The training pipeline is shown in the Algorithm. \ref{alg: curriculum learning}.}

\begin{algorithm}[t]
\footnotesize
    \caption{Iterative Curriculum Learning.}
    \label{alg: curriculum learning}
    \begin{algorithmic}[1]
    \REQUIRE The chatbot $\mathcal{C}_{\theta}$ with parameter $\theta$. User set $\{\mathcal{U}_i\}_{i=1}^N$. The maximum number of iterations $K$ and dialogue turns $T$. 
    \FOR{$k = 1, \dots, K$}
        \STATE Collect a corpus $\{\mathcal{D}_i\}_{i=1}^N$ according to Algorithm \ref{alg:datacollect_algorithm}.
        \STATE Initialize $\mathcal{D}^* = \varnothing$.
        \FOR{$i = 1, \dots, N$}
            \STATE $is\_easy \gets \TRUE$
            \FOR{$t = 1, \dots, T$}
                \STATE Calculate ${Dif}(u_{i,t}^{Bot})$ via Eqn. (\ref{equation: diff}).
                \IF{${Dif}(u_{i,t}^{Bot})$ == \TRUE}
                    \STATE $is\_easy \gets \FALSE$
                    \STATE \textbf{break}
                \ENDIF
            \ENDFOR
            \STATE Add $\mathcal{D}_i$ to $\mathcal{D}^*$ \textbf{if} $is\_easy$.
        \ENDFOR
    \STATE Fine-tuning the model $\mathcal{C}_{\theta_k}$ with $\mathcal{D}^*$. 
    \ENDFOR
    \RETURN Chatbot $\mathcal{C}_{\theta_K}$.
    \end{algorithmic}
    \vspace{-2pt}
\end{algorithm}

\subsection{User Background Dataset~\textit{ISCO-800}}
\label{sec:isco800}

\c{To train a chatbot capable of interacting with users from diverse backgrounds}, we construct a dataset called \textit{ISCO-800}, containing 800 pieces of user background information.  
Specifically, to ensure the dataset is representative of global occupations, we refer to the ISCO-08 classification by the International Labour Organization~\cite{ISCO}.  We randomly select 40 out of 43 sub-major occupation groups from ISCO-08 as the source data. The details of these occupation groups are listed in Appendix C.
We designed a prompt in Appendix C to guide GPT-4 in generating realistic user backgrounds for each occupation. These backgrounds include names, career histories, educations, personalities, and hobbies. Each user's background consists of 50 to 100 words. To ensure realism, not all users have positive personalities or smooth careers. We generate 20 different backgrounds for each occupation group, resulting in 800 user backgrounds. 
Statistics of \textit{ISCO-800} and comparison with other related datasets in terms of the user background are in Appendix C.

\section{Experiment}
\subsection{Experimental Settings}

\textbf{Implementation Details.}
\c{We implemented~\methodname with Qwen1.5-32B-Chat~\cite{qwen} as the base LLM. We use user backgrounds from \textit{ISCO-800} to form the user agents and each engages in 5-turn dialogues with the chatbot. The 800 user agents are divided into training, validation, and test sets (500, 100, and 200 users, respectively) for dialogue generation. They are not duplicated in the occupation groups. \cam{This ensures the model is always evaluated on unseen domains during testing}. See Appendix B and C for more details.}

\textbf{Compared Methods.} 
We compare~\methodname with existing open-domain dialogue methods, including the universal LLMs,  recent target-oriented open-domain dialogue methods (ProCoT~\cite{deng2023prompting} and TRIP~\cite{wang2024target}) and other LLM-based methods (BlenderBot 3~\cite{shuster2022blenderbot}, MemoChat~\cite{lu2023memochat} and MemoryBank~\cite{zhong2024memorybank}). For a fair comparison, we adapt them to the user-oriented dialogue tasks by fine-tuning the models or using prompt strategies. See Appendix B for details.

\textbf{Metrics.} 
\hjw{We use our designed scoring system (including the background relevance $Rel.$,  user's interest level $Int.$, and response value $Val.$) to evaluate the chatbot's user-oriented proactivity. Notably, the critic used for evaluation includes gpt-3.5-turbo-0125 or gpt-4-turbo-2024-04-09.
In addition, we also adopt the widely used perplexity (PPL)~\cite{jelinek1977perplexity} to measure the quality of the generated corpus.}

\textbf{Evaluation.} 
In the test phase, we obtain the user backgrounds of the test set from \textit{ISCO-800} and construct different user agents. Then we let the chatbot start 5 turns of open-domain dialogue with each user agent. The chatbot does not obtain the user's background information in advance. We use the critic to evaluate metrics of user-oriented proactivity of each turn of the chatbot's response for the generated dialogue corpus and obtain the averaged scores.

\begin{table*}[t]
\vspace{-9pt}
\small

  \centering 
  \scalebox{0.92}{
  \renewcommand{\tabcolsep}{6.4pt}
  \hspace*{-0.5cm}
  \begin{tabular}{ccccccccccc}  
    \toprule
    
    \multirow{2}{*}{Category} & \multirow{2}{*}{Methods} & \multirow{2}{*}{Param.} & \multicolumn{3}{c}{Critic: GPT-3.5}   & \multicolumn{3}{c}{Critic: GPT-4} & \multirow{2}{*}{PPL $\downarrow$}   \\

    & & & $Rel.\uparrow$ & $Int.\uparrow$& $Val.\uparrow$ &  $Rel.\uparrow$ & $Int.\uparrow$& $Val. \uparrow$  \\
  
    \midrule 

    \multirow{6}{*}{\makecell{Universal\\LLMs}} & Llama-3-70B-Instruct \cite{llama3modelcard}   & 70B       & 4.702 & 3.776 & 3.773 & 4.619  & 3.628  & 3.734 & 10.153\\
                                                  & Qwen1.5-72B-Chat \cite{qwen}                & 72B       & 4.553 & 3.670 & 3.664 & 4.752  & 3.672  & 3.764 & 11.263\\
                                                  & Qwen1.5-32B-Chat \cite{qwen}                & 32B       & 4.534 & 3.672 & 3.639 & 4.690  & 3.640 & 3.746 & 11.038\\
                                                  & Vicuna-33b-v1.3 \cite{chiang2023vicuna}     & 33B       & 4.508 & 3.626 & 3.609 & 4.507  & 3.595  & 3.644 & 10.821\\
                                                  & GLM-4 \cite{glm4}                          & NA       & 4.401 & 3.581 & 3.486 & 4.640  & 3.657  & 3.668 & 12.326 \\
                                                  & GPT-3.5 Turbo \cite{openai_chatgpt}         & $\approx$175B     & 4.625 & 3.719 & 3.701 & 4.760  & 3.579  & 3.688 & 12.691\\
                                                  & GPT-4o \cite{GPT4o}         & NA     & 4.591 & 3.546 & 3.578 & 4.599  & 3.656  & 3.644 & 17.624\\
  \midrule                                                                 
  \multirow{2}{*}{\makecell{Target-Oriented \\ Methods}} & ProCoT~\cite{deng2023prompting}    & $\approx$175B  & 4.363  & 3.526 & 3.496 & 4.279   & 3.001   & 3.099  & 19.895 \\
                                                                   & TRIP~\cite{wang2024target}    & 32B  & 4.521  & 3.634 & 3.626 &  4.341   & 3.461 & 3.455 & 9.327 \\                                               
    \midrule 
    \multirow{4}{*}{\makecell{Other LLM-Based \\ Methods}}      & BlenderBot 3 \cite{shuster2022blenderbot}  & 30B      & 3.414 & 3.188 & 3.056 & 3.231 & 3.124 & 3.008 & 13.287 \\
                                                                   & MemoChat \cite{lu2023memochat}          & 33B      & 4.609  & 3.713  & 3.771  & 4.203  & 3.438  & 3.619 & 8.773  \\
                                                                   & MemoryBank~\cite{zhong2024memorybank}    & 6B       & 3.798 & 3.220 & 3.253 &3.312  & 2.676 & 2.996 & 12.564 \\

\rowcolor{pink!30}
                                                                   & \textbf{Ours~(Qwen1.5-32B-Chat)}   & 32B    & \textbf{4.858} & \textbf{3.943}& \textbf{3.925}  & \textbf{4.818} & \textbf{3.721}& \textbf{3.903} & \textbf{7.956}\\

    \bottomrule
  \end{tabular}
  }
  \vspace{-5pt} 
  \caption{\hjw{Comparsion experimental results on the \textit{ISCO-800}.}}\label{tab:comparation}
  \vspace{-5pt} 
\end{table*}

\subsection{Comparison Experiments}
\label{subsec: compare}

 We compare our \methodname~with other open-domain dialogue methods and the results are shown in Table~\ref{tab:comparation}. Our proposed \methodname, with only 32B parameters, outperforms all other methods in terms of $Rel.$, $Int.$, $Val.$, and PPL, demonstrating its effectiveness. The detailed analyses are as follows.

\textbf{\methodname~outperforms universal LLMs with fewer parameters.} 
With GPT-3.5 as the critic, our \methodname~outperforms the strongest baseline Llama-3-70B-Instruct by 3.32\%, 4.42\%, and 4.03\% on $Rel.$, $Int.$ and $Val.$. With GPT-4 as the critic, our \methodname~outperforms the strongest baseline Qwen1.5-72B-Chat by 1.39\%, 1.33\% and 3.69\% on $Rel.$, $Int.$ and $Val.$. Our method also reduces the PPL by 21.64\% compared with Llama-3-70B-Instruct and 29.36\% compared with Qwen1.5-72B-Chat. It indicates that our \methodname~improves the performance even with half of the model parameters. We attribute the improvement to the critic-guided corpus construction paradigm that facilitates the LLM to generate user-interested dialogue data. Besides, the iterative curriculum learning method helps adapt to different users from easy to hard, enhancing the chatbot's user-oriented proactivity. Moreover, the GPT-4-based critic exhibits similar scoring trends to the GPT-3.5-based critic. It suggests that although we use GPT-3.5-based critic during training, we still obtain a consistent judgment with the better-performing GPT-4 in the test. Therefore, we use GPT-3.5-based critic in ablation studies.

\textbf{\methodname~outperforms the target-oriented methods}. \methodname~outperforms the strongest baseline TRIP by 7.45\%, 8.50\%, and 8.25\% on $Rel.$, $Int.$ and $Val.$ (Critic: GPT-3.5) and 10.99\%, 7.51\%, and 12.97\% (Critic: GPT-4). \methodname~also reduces the PPL by 14.70\%. We attribute this to the fact that although it has proactivity in target topic switching, the topic is not necessarily the one that the user is interested in, leading to lower performance of user-oriented proactivity. 

\textbf{\methodname~outperforms other LLM-based methods}.~\methodname~outperforms the strongest MemoChat  by 5.40\%, 6.19\%, and 4.08\% on $Rel.$, $Int.$ and $Val.$ (Critic: GPT-3.5) and 14.63\%, 8.23\%, and 7.85\% (Critic: GPT-4). Our method also reduces the PPL by 9.31\%. We attribute this to the reason that MemoChat’s focus on maintaining consistency offers limited improvement in open-domain conversations where topics shift constantly, whereas our trial-and-error strategy yields better performance across a diverse range of topics.

\begin{table}[t]
\small
  
  \centering
  \scalebox{0.9}{
  \renewcommand{\tabcolsep}{9.0pt}
  \begin{tabular}{cccc|ccccc}
    \toprule
    SFT & CDC & IFT & CL & $Rel.\uparrow$ & $Int.\uparrow$ & $Val.\uparrow$  \\\midrule
    & & &                                           & 4.522 & 3.629 & 3.620   \\

\usym{1F5F8} & & &                                           & 4.423 & 3.547 & 3.525   \\
\usym{1F5F8} &\usym{1F5F8}  &&                               & 4.534 & 3.672 & 3.639   \\
 &\usym{1F5F8} & \usym{1F5F8} &                  & 4.578 & 3.738 & 3.758   \\
\rowcolor{pink!30}
 &\usym{1F5F8} & \usym{1F5F8} &   \usym{1F5F8}   & \textbf{4.858} & \textbf{3.943}& \textbf{3.925} \\
    \bottomrule
\end{tabular}
}
\caption{Ablation experiments on \textit{ISCO-800}. The row without ticks means the original Qwen1.5-32B-Chat.}\label{tab:ablation_training_strategies}
\vspace{-8pt}
\end{table}

\subsection{Ablation Studies}
\label{subsec: ablation}

We conducted ablation experiments with the GPT-3.5-based critic to validate the
effectiveness of our proposed iterative curriculum learning. We compared different training strategies in our iterative curriculum learning, described as follows:

\textbf{SFT}: An initial LLM generates a dialogue corpus through multiple dialogues with each user agent for supervised fine-tuning, without the critic's involvement.
\textbf{SFT+CDC (Critic-guided Dialogue Corpus Collection)}: Based on SFT, a critic evaluates the chatbot's responses and requests regenerations for low-rated responses, producing a high-quality corpus for fine-tuning.
\c{We repeat the above two strategies for the same number of iterations as~\methodname to ensure a consistent amount of training data.}
\textbf{CDC+IFT (Iterative Fine-Tuning)}: The chatbot engages in dialogue with each user agent, incorporating critic feedback. The collected corpus is used for initial fine-tuning. Then the fine-tuned chatbot re-engages with each user agent and collects the corpus. This process of corpus collection and fine-tuning \c{undergoes the same iterative rounds as~\methodname}, gradually enhancing the performance of the chatbot.
\textbf{CDC+IFT+CL (Curriculum Learning)}: Building on CDC+IFT, we additionally use our Communication Difficulty Measurer to identify easy samples after each dialogue generation, and we only use easy samples for fine-tuning.

\hjw{\textbf{Effectiveness of CDC.} 
As in Table \ref{tab:ablation_training_strategies}, \methodname~(SFT+CDC) outperforms \methodname~(SFT) in all metrics, as the critic feedback improves the dialogue corpus quality, resulting in better fine-tuning and enhanced LLM performance \cite{zhou2024lima}.} Besides, \methodname~(SFT) performs even inferior to that of the original LLM. We attribute this to the poor-quality corpus generated during training data collection without feedback.

\hjw{\textbf{Effectiveness of IFT.}
As in Table \ref{tab:ablation_training_strategies}, compared to only CDC, training with CDC and IFT improves the user-oriented proactivity of chatbots by 0.97\%, 1.80\%, and 3.27\% on $Rel.$, $Int.$, and $Val.$, respectively. It indicates that re-collecting the dialogue corpus data using the fine-tuned model helps to improve the quality of the generated data and improves the performance of the model in the next round of fine-tuning.}

\hjw{\textbf{Effectiveness of CL.} As in Table \ref{tab:ablation_training_strategies},~\methodname~ has significant improvement compared to \methodname~ (w/o CL). Specifically, there are 6.12\%, 5.48\% and 4.44\% increases on $Rel.$, $Int.$, and $Val.$ respectively. In conclusion, curriculum learning helps the model learn the essential features of the task more quickly and improves the generalization ability~\cite{wang2021survey}.}

\begin{figure}[t]
        \centering
        \vspace{-5pt}
        \includegraphics[width=0.95\linewidth]{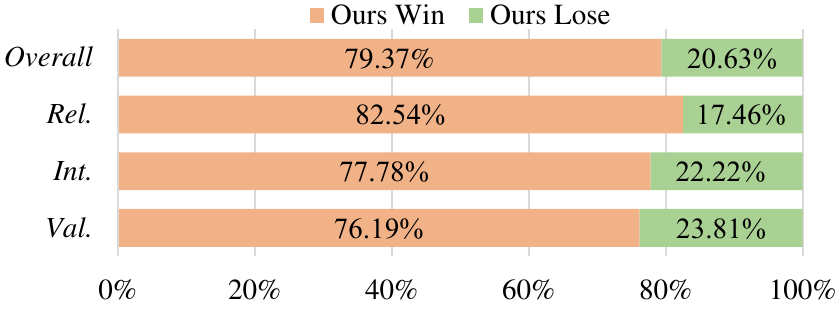}
        \vspace{-5pt}
        \caption{\c{Real user evaluation results between our~\methodname and Llama-3-70B-Instruct baseline.}}
        \vspace{-10pt}
        \label{fig:human_eval}
\end{figure}

\subsection{More Discussions}
\label{sub: more_discussion}

\hjw{\textbf{Online Chatbot Interaction with Real Users.} To illustrate the practicality of~\methodname, we deploy~\methodname and one of the strongest baselines Llama-3-70B-Instruct on the Internet and recruited 62 participants to have free chat with each of the two chatbots. Participants are unknown to the names of the chatbots. The participants are first advised to chat with the two chatbots separately with any identity and topics. Then they are required to compare the performance of the two chatbots.
See Appendix D for detailed settings. \c{As in Figure \ref{fig:human_eval}, over 75\% of participants preferred~\methodname in the overall performance as well as the $Rel.$, $Int.$ and $Val.$. 
It indicates our \methodname has enhanced performance and aligns with human preferences. Most Participants consider that~\methodname pays more attention to their background and chatting interests, gives more valuable responses, and leads to a better chatting performance. It reveals the practical value in real chatting scenarios.}}

\begin{figure}[t]
        \vspace{-9pt}
        \centering
        \includegraphics[width=0.95\linewidth]{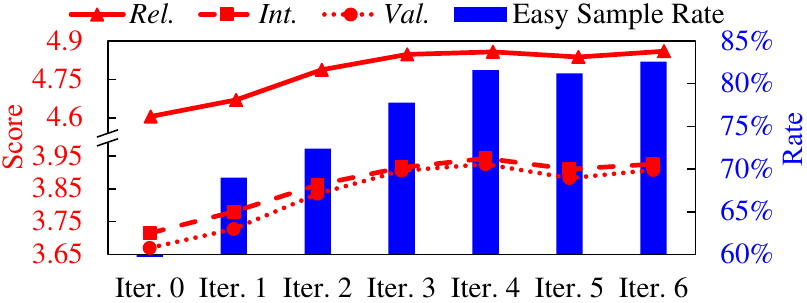}
        \vspace{-8pt}
        \caption{The performance during the iterative curriculum learning. The curves represent the metrics of user-oriented proactivity. The bars represent the rate of the easy sample.}
        \vspace{-2pt}
        \label{fig:curves}
\end{figure}

\textbf{Performance evolution during iterative curriculum learning.} 
To illustrate the performance improvement process during iterative curriculum learning, we present our designed metrics on the validation set and the rate of easy samples in the training data at each iteration in Figure \ref{fig:curves}. It shows that performance and the rate of easy samples gradually improve with each iteration, converging at the $4^{th}$ iteration. This suggests that iterative curriculum learning progressively enhances model performance by training the chatbot to adapt to users with increasing communication difficulty. Additionally, the maximum number of iterations is set to 4 due to convergence. We further calculate the regeneration rate and average number of regenerations during the training process. As in Table \ref{tab:icl}, both the regeneration rate and average regenerations decrease as the iteration proceeds. It indicates that through iterative curriculum learning, the cost of regeneration is reduced as the iteration progresses. Besides, the chatbot increasingly delivers satisfactory responses on the first attempt.

\begin{table}[t]
\small

  
  \centering
  \scalebox{0.9}{
  \renewcommand{\tabcolsep}{9.0pt}
  \begin{tabular}{ccccc}
    \toprule
    Training Process & Iter. 1  & Iter. 2 & Iter. 3  & Iter. 4  \\
    \midrule
    \makecell{Regeneration Rate}  & 37.7\% & 37.9\% & 26.0\%  & 23.2\%  \\
    \makecell{Average Number of \\ Regenerations}  & 0.785 & 0.784 & 0.552 & 0.474  \\

    \bottomrule
\end{tabular}
}
\vspace{-5pt}
\caption{Regeneration rate and average number of regenerations during the iterative curriculum learning.}\label{tab:icl}
\vspace{-5pt}
\end{table}

\textbf{Effects of LLM category and size.} To verify the applicability of~\methodname, we conduct experiments on LLMs of different categories and sizes. In Table~\ref{tab:ablation_llms}, for various sizes in Qwen1.5 series, ~\methodname consistently improves performance, with gains of 5\%$\sim$10\%. For LLMs of similar size but different types, e.g., Qwen1.5 and Vicuna, \methodname~also yields improvements. Besides, stronger base LLMs like original Qwen1.5-14B and Qwen1.5-32B show greater gains than Vicuna-33B. In summary, \methodname~works well across LLM categories and sizes, and brings larger gains on stronger base LLMs.

\subsection{Case Study}  

\begin{figure}[t]
\vspace{-9pt}
    \centering
    \includegraphics[width=0.98\linewidth]{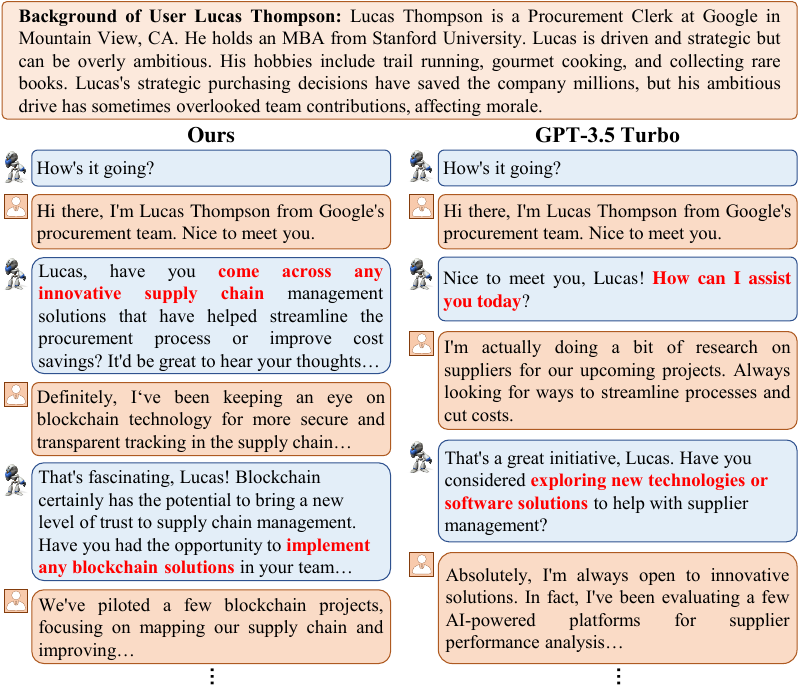}
    \vspace{-8pt}
        \caption{Case study of our \methodname~compared with GPT-3.5-Turbo.}
        \label{fig:case_study_1}
    \vspace{-5pt}
\end{figure}

\begin{table}[t]
\small 
\renewcommand{\tabcolsep}{6pt}

  
  \centering
  \scalebox{0.9}{
  \renewcommand{\tabcolsep}{13.0pt}
  \begin{tabular}{cccc}
    \toprule
    LLMs & $Rel.\uparrow$ & $Int.\uparrow$ & $Val.\uparrow$      \\
    \midrule 
    Original Qwen1.5-14B-Chat                  & 4.356 & 3.550 & 3.490  \\
\rowcolor{pink!30}    Ours (Qwen1.5-14B-Chat)            & 4.653 & 3.811 & 3.789    \\
    Original Vicuna-33B-Chat                 & 4.323 & 3.514 & 3.479   \\
\rowcolor{pink!30}    Ours (Vicuna-33B-Chat)           & 4.363 & 3.568 & 3.479  \\
    Original Qwen1.5-32B-Chat                & 4.522 & 3.629 & 3.620  \\
\rowcolor{pink!30}    Ours (Qwen1.5-32B-Chat)        & \textbf{4.858} & \textbf{3.943}& \textbf{3.925}   \\
    \bottomrule
  \end{tabular}
  }
  \vspace{-5pt}
  \caption{Results of different size and types of LLMs on \textit{ISCO-800}.}\label{tab:ablation_llms}
  \vspace{-7pt}
\end{table}

In Figure \ref{fig:case_study_1}, we provide test set samples to compare the performance of  \methodname~with GPT-3.5-Turbo baseline. After the introduction of the user, \methodname~ask specific questions about the user’s background, like ``Innovative supply chain management solutions''. When the user mentions blockchain technology, \methodname~asks about its implementation. The responses are highly interactive and relevant to the user's interests, keeping the user engaged in the discussion.
In contrast, the dialogues of GPT-3.5 are more generic, with the chatbot relying on simple pleasantries like ``How can I assist you today?'' and failing to address the user’s specific background. 
The responses are shallow, do not effectively explore the user’s interests, and result in superficial dialogues that neither engage the user nor encourage further discussion. Thus, our \methodname~exhibits better user-oriented proactivity. Another case between our \methodname~and target-oriented method TRIP is provided in Appendix F.

\section{Conclusion}
We propose a User-oriented Proactive Chatbot to address the lack of user-oriented proactivity in open-domain dialogue. We first construct a critic to evaluate the user-oriented proactivity. Then we use the critic to guide dialogues between the chatbot and user agents, generating a corpus with enhanced user-oriented proactivity. We introduce a user background dataset \textit{ISCO-800} to ensure the diversity of user backgrounds. Finally, we train the chatbot with an iterative curriculum learning strategy to adapt to different users from easy to hard. Experiments demonstrate that our proposed training method is applicable to different LLMs, improving user-oriented proactivity in open-domain dialogues. We hope to bring new insight into the improvement of dialogue quality for LLMs and human-computer interaction experience.

\section*{Acknowledgments}
This work was supported in part by the National Key R\&D Program of China under Grant 2022YFB4500600, in part by the Guangdong Provincial Key Laboratory of Human Digital Twin under Grant 2022B1212010004, and in part by Major Key Project of Peng Cheng Laboratory (PCL) PCL2023A08.

\appendix




\bibliographystyle{named}
\bibliography{reference}

\end{document}